\documentclass{tlpmod}

\usepackage[T1]{fontenc}

\usepackage{mytheorems}
\newcommand{\openbox}{\leavevmode
  \hbox to.77778em{%
  \hfil\vrule
  \vbox to.675em{\hrule width.6em\vfil\hrule}%
  \vrule\hfil}}

\usepackage[fleqn]{mathtools}
\usepackage{hyperref}

\usepackage{iftex}
\usepackage{wrapfig}
\usepackage{graphicx}
\usepackage{float}
\usepackage{booktabs}
\usepackage{placeins}
\usepackage{algorithm}
\usepackage{algpseudocode}
\usepackage{xcolor}
\usepackage{tcolorbox}
\usepackage{subcaption}
\usepackage{multirow}
\usepackage{url}
\usepackage{natbib}
\setcitestyle{numbers,square,comma} 
\usepackage{alltt}

\usepackage{pseudo}
\pseudoset{
ct-left={{~\small//~}},
ct-right=\relax,
indent-length=1em,
label={\small\arabic*},
kw,
}

\DeclareMathAlphabet{\mathcal}{OMS}{cmsy}{m}{n}
\SetMathAlphabet{\mathcal}{bold}{OMS}{cmsy}{b}{n}

\ifpdf
  \usepackage[strings]{underscore}         
  \usepackage[T1]{fontenc}        
\else
  \usepackage{breakurl}           
\fi

\usepackage{xcolor}

\definecolor{DodgerUniformBlue}{rgb}{0.0,0.353,0.612}
\newcommand{\define}[1]{\emph{\textcolor{DodgerUniformBlue}{#1}}}

\def\codefont{\upshape\ttfamily}
\def\code#1{\text{\codefont #1}}

\newcommand{\CM}{\mathbin{\code{\small :--}}}
\newsavebox{\notBox}
\begin{lrbox}{\notBox}\verb|not|\end{lrbox}
\newcommand{\NOT}{\mathop{\usebox{\notBox}}}

\begin{document}



\jnlPage{1}{14}
\jnlDoiYr{2024}
\doival{10.1017/xxxxx}

\title{CON-FOLD\\ Explainable Machine Learning with Confidence}

\begin{authgrp}
\author{\sn{McGinness} \gn{Lachlan}}
\affiliation{School of Computer Science, ANU, and CSIRO/Data61, Canberra, Australia}
\author{\sn{Baumgartner} \gn{Peter}}
\affiliation{CSIRO/Data61, and School of Computer Science, ANU, Canberra, Australia}
\end{authgrp}

\maketitle

\begin{abstract}
        FOLD-RM is an explainable machine learning classification algorithm that uses training data to create a set of classification rules.
        In this paper we introduce CON-FOLD which extends FOLD-RM in several ways.
        CON-FOLD assigns probability-based confidence scores to rules learned for a classification task. This allows users to know how confident they should be in a prediction made by the model. We present a confidence-based pruning algorithm that uses the unique structure of FOLD-RM rules to efficiently prune rules and prevent overfitting. 
        Furthermore, CON-FOLD enables the user to provide pre-existing knowledge in the form of logic program rules that are either (fixed) background knowledge or (modifiable) initial rule candidates.
        The paper describes our method in detail and reports on practical experiments. We demonstrate the performance of the algorithm on benchmark datasets from the UCI Machine Learning Repository. For that, we introduce a new metric, Inverse Brier Score, to evaluate the accuracy of the produced confidence scores.
        Finally we apply this extension to a real world example that requires explainability: marking of student responses to a short answer question from the Australian Physics Olympiad. 
        \end{abstract}

\section{Introduction}

Machine Learning (ML) has been shown to be incredibly successful at learning patterns from data to solve problems and automate tasks. However it is often difficult to interpret and explain the results obtained from ML models. 
Decision trees are one of few ML methods that offer transparency with regards to how decisions are made. They allow users to follow a set of rules to determine what the outcome of a task (say classification) should be. The difficulty with this approach is finding an algorithm that is able to construct a reliable set of decision trees.

One approach of generating a set of rules equivalent to a decision tree is the First Order Learner of Default (FOLD) approach introduced by Shakerin et.\ al.\ in 2017 \cite{Shakerin2017New}. 
To improve a model's ability to handle exceptions in rule sets, Shakerin, Wang, Gupta and others introduced and refined an explainable ML algorithm called  First Order Learner of Default (FOLD) \cite{Wang2022Explainable, Shakerin2017New, Wang2022FOLDR++, Wang2022FOLD-RM, Wang2023FOLD-SE, Parth24NeSyFOLD} which learns non-monotonic stratified logic programs \cite{Quinlan1990Learning}. The FOLD algorithm is capable of handling numerical data (FOLD-R) \cite{Shakerin2017New}, multi-class classification (FOLD-RM) \cite{Wang2022FOLD-RM} and image inputs (NeSyFOLD) \cite{Parth24NeSyFOLD}. FOLD-SE uses Gini Impurity instead of information gain in order to obtain a more concise sets of rules \cite{Wang2023FOLD-SE}. Thanks to these improvements, variants of the FOLD algorithm are now competitive with state of the art ML techniques such as XGBoost and RIPPER in some domains \cite{Wang2022FOLDR++, Wang2023FOLD-SE}. 

The rules produced by the FOLD algorithm are highly interpretable, however they can be misleading. As an example let's consider the popular Titanic dataset~\cite{titanic}, where passengers are classified into to two categories: perished or survived. One rule from the FOLD algorithm might say that a passenger survives if they are female and do not have a third class ticket. When given a new set of data and this rule, a user might be unpleasantly surprised to find that such a passenger perished. This is because the rule can have the appearance of being definitive. In reality this could be a good rule that is correct 99\% of the time. In order for a FOLD model to be more understandable and trustworthy for users, a confidence value could be provided. This would provide the user with a measure of the certainty of the rule and would make it clear to the user that not all women with second class tickets survive.

In this paper we introduce Confidence-FOLD (CON-FOLD), an extension of the FOLD-RM
algorithm. CON-FOLD provides confidence values for each rule generated by the FOLD
algorithm so users know how confident they should be for each rule in the model. In
addition we present a pruning algorithm that makes use of these confidence values.
These techniques applied to the Titanic example yield the following rules and
confidences:\footnote{For simplicity of demonstration, the rules were obtained from the
  Titanic \emph{test} dataset.
We used an improvement threshold of $0.1$ (Section~\ref{sec:improvement-threshold-pruning})
and a confidence threshold of $0.5$ (Section~\ref{sec:confidence-threshold-pruning}).
}
\begin{alltt}
\quad survived(X, false) :- rule1(X).                 %confidence: 0.9
\quad survived(X, true) :- rule2(X), not rule1(X).    %confidence: 0.97
\quad rule1(X) :- not sex(X, female). 
\quad rule2(X) :- sex(X, female). 
\end{alltt}

We also provide a metric, Inverse Brier Score, which can be used to evaluate probabilistic or confidence-based predictions that maintains compatibility with the traditional metric of accuracy. We provide capability to introduce modifiable initial knowledge into a FOLD model. Finally we demonstrate the effectiveness of adding background knowledge in the marking of student responses to physics questions. Note that we choose to focus on multi-class classification tasks our experiments, however CON-FOLD can also be applied to binary classification.

\section{Formal Framework and Background}
\label{sec:LearningFramework}
We work with usual logic programming terminology.
A \define{(logic program) rule} is of the form  
    \begin{equation}
      \label{eq:rule}
      h \CM l_1,\ldots,l_k, \NOT e_1, \ldots ,\NOT e_n \enspace.
    \end{equation}
where the $h$, $l_1,\ldots,l_k$ and $e_1,\ldots,e_n$ all are atoms, for some $k, n \ge 0$. 
A \define{program} is a finite set of rules. 
We adopt the formal learning framework described for the FOLD-RM algorithm. 
In this, two-ary predicate
symbols are used for representing feature values, which can be categorial or
numeric. Example feature atoms are \verb|name(i1,sam)| and \verb|age(i1,30)| of an individual \code{i1}.  
Auxiliary
predicates and Prolog-like built-in predicates can be used as well. 
Rules always pertain to one single individual and its features as in this example:
\begin{alltt}
\quad female(X) :- rule1(X).\hfill\textnormal{(1)}
\quad rule1(X) :- age(X,A), A>16, not ab(X). \hfill\textnormal{(2)}
\quad ab(X) :- name(X,sam), not fav_color(X,purple). \hfill\textnormal{(3)}
\end{alltt}
These rules for a target relation \verb|female| could have been learnt from training data where all individuals
older than 16 are female, except those named \verb|sam| and whose
favourite color is not \verb|purple|. Below we use the letter $r$ to refer to the rule for
the target relation, $r = $~\verb|female|, (1) in this example, and the letter $R$ for the set of auxiliary 
rules, $R = \{\textnormal{(2)}, \textnormal{(3)}\}$,  that are needed to define the predicates in $r$.

The learning task in general is defined in~\cite{Wang2022FOLD-RM}.
The learning algorithm takes as input two disjoint sets $X = X_p \uplus X_n$ of
\define{positive and negative training examples}, respectively. It assumes that the training set distribution is approximately the same as the test
set distribution.

For any $d \in X$, let 
\define{$\id{features}(d)$} denote $d$'s features as a set  of atoms
over some a priori fixed Skolem constant, say, \code{c}. Example: 
$\id{features}(d) = \{\code{age(c,18)}, \code{name(c,adam)}, 
\code{fav_color(c,red)}\}$. The learning algorithm below checks if a current set $R \cup
\{r\}$ puts an example $d$ into the target class. Letting \define{$\ell$} denote the target
class as an atom, e.g., $\ell = \code{female(c)}$, this is accomplished via an entailment
check $R \cup \{r\} \cup \id{features}(d) \models \ell$.

Because all learned programs are stratified, we adopt the standard
semantics for that case, the perfect model semantics (layered bottom-up fixpoint
computation), which is also reflected in the FOLD-RM algorithm. We emphasize that default
negation causes no problems for calculating confidence scores of a rule.  

The confidence
scores are attached only to the top-level rules $r$, never to the rules $R$ referred to under default
negation. That is, having to deal with confidence scores in negated context will never be necessary.
We will describe the rationale behind this design decision shortly below.

The Boolean learning task is to determine a stratified program $P$ such that   
\begin{align*}
P \cup \id{features}(d) \models \ell & \text{ for all $d \in X_p$} & \text{and} & 
&
                           P \cup \id{features}(d) \not\models \ell & \text{ for all $d \in X_n$}\text{ .} 
\end{align*}
The multi-class learning problem is a generalization to a finite set of classes. 
The training data $X$ then is comprised of atoms indicating the target class, for each
data point, e.g.\ \code{survived(c,false)} or \code{survived(c,true)} in the Titanic example.
Conversely, by splitting multi-class learning problems can be treated as a sequence of Boolean learning problems.

The basis of CON-FOLD is the FOLD-RM algorithm~\cite{Wang2022FOLD-RM}. 
FOLD-RM simplifies a multi-class classification task into a series of Boolean
classification task. It first chooses the class with the most examples and sets data which
correspond to this class as positive and all other classes as negative. It then generates
a rule that uses input features to maximise the information gain. This process is
repeated on a subset of the training examples by eliminating those that are correctly
classified with the new rule. The process stops when when all examples are classified.

\section{Related Work}

Confidence scores have been introduced in decision tree learning for assessing the admissibility of a
pruning step~\cite{quinlan1990}. In essence, decision tree pruning removes a sub-tree if
the classification accuracy of the resulting tree does not fall below a certain threshold,
e.g., in terms of standard errors, and possibly corrected for small domain sizes.
Decision trees can be expressed as sets of production rules~\cite{quinlansimplifying1987},
one production rule for each branch in a decision tree. The production rules can again be
simplified using scoring functions~\cite{quinlansimplifying1987}.

The FOLD family of algorithms learns rules with exceptions by means of default negation
(negation-as-failure). The rules defining the exceptions can have exceptions
themselves. This sets FOLD apart from the production systems learned by decision tree
classifiers, which do not take advantage of default negation. This may lead to more complex rule sets.
Indeed, \cite{Wang2022FOLD-RM} observe
that ``For most datasets we experimented with, the number of leaf nodes in the trained C4.5
decision tree is much more than the number of rules that FOLD-R++/FOLD-RM generate. The
FOLD-RM algorithm outperforms the above methods in efficiency and scalability due to (i)
its use of learning defaults, exceptions to defaults, exceptions to exceptions, and so on
(i) its top-down nature, and (iii) its use of improved method (prefix sum) for heuristic
calculation.'' We do feel, however, that these experimental results could be completed from a
more conceptual point of view. This is beyond the scope of this paper and
left as future work. 

Scoring functions have been used in many rule-learning systems. The common idea is to
allow accurracy to degrade within given thresholds for the benefit of simpler rules.
See~\cite{DBLP:journals/corr/abs-2005-00904} for a discussion of the more recent ILASP3
system and the references therein.  Hence, we do not claim originality of using scoring
systems for rule learning.  We see our main contribution differently and not in
competition with other systems.  Indeed, one of the main goals of this paper is to equip
an \emph{existing} technique that has been shown to work well -- FOLD-RM -- with
confidence scores. With our algorithm design and experimental evaluation we show that this
goal can be achieved in an ``almost modular'' way. Moreover, as our extension requires only
minimal changes to the base algorithm, we expect that our method is transferrable to other
rule learning algorithms.

\section{The CON-FOLD Algorithm and Confidence Scores}
\label{sec:CON-FOLD}
The CON-FOLD algorithm assigns each rule a confidence score as it is created. For easy
interpretability, confidence scores should be equal to probability values (p-values from a
binomial distribution) in the case of large amounts of data. However, p-values would
be a very poor approximation in the case of small amounts of training data; if a rule
covered only one training example it would receive a confidence value of 1
($100\%$). There are many techniques for estimating p-values from a sample, we chose the
centre of the Wilson Score Interval \cite{Wilson1927} given by Equation
\ref{eqn:confidence} below.  The Wilson Score Interval adjusts for the asymmetry in the binomial
distribution which is particularly pronounced in the case of extreme probabilities and
small sample sizes. It is less prone to producing misleading results in these situations
compared to the normal approximation method, thus making it more trustworthy for users
\cite{Agresti1998Approximate}.
\begin{equation}
    p = \frac{n_p + \frac{1}{2} Z^2}{n + Z^2}\text{, }
    \label{eqn:confidence}
\end{equation}
where $p$ is the confidence score, $n_p$ is the number of training examples corresponding to the target class covered by the rule, $n$ is the number of training examples covered by the rule corresponding to all classes, and $Z$ is the standard normal interval half width, by default we use $Z=3$.

\begin{theorem}
In the limit where there is a large amount of data classified by a rule ($n \rightarrow \infty$), the confidence score approaches the true probability of the sample being from the target class.
\end{theorem}

\begin{proof}
The true probability that a randomly selected example that follows the provided rule is
from the target class is $p_r=\frac{n_p}{n}$.
As $n$ increases, the law of large numbers states that the portion of examples which belong to the target class becomes
$\lim_{n \rightarrow \infty} n_p = p_r n$.
Also, as $n \rightarrow \infty$, the relative contribution of $Z^2$ terms becomes negligible. Therefore:
\[
    \lim_{n \rightarrow \infty} p = \lim_{n \rightarrow \infty} \frac{n_p + \frac{1}{2} Z^2}{n + Z^2} = \lim_{n \rightarrow \infty}
    \frac{n p_r}{n} = p_r \hspace*{8em}\text{\qed}
\]
\end{proof}

Once rules have the associated confidence scores they are expressed as follows:
    \begin{equation}
      \label{eq:ProbRule}
     p:: h \CM l_1,\ldots,l_k, \NOT e_1, \ldots ,\NOT e_n \enspace.
    \end{equation}
    where $p$ is the confidence score.
The format of confidence score annotations is directly supported by 
probabilistic logic programming systems such as Problog~\cite{Problog} and
Fusemate~\cite{Baumgartner2023Bottom}. In this paper, we do not explore this possibility
and just let the confidence scores allow the user to know the reliability of a
prediction made by a rule in the logic program. 
\vspace{0.25cm}

\begin{figure*}[ht]
\centering
\includegraphics[width=0.9\textwidth]{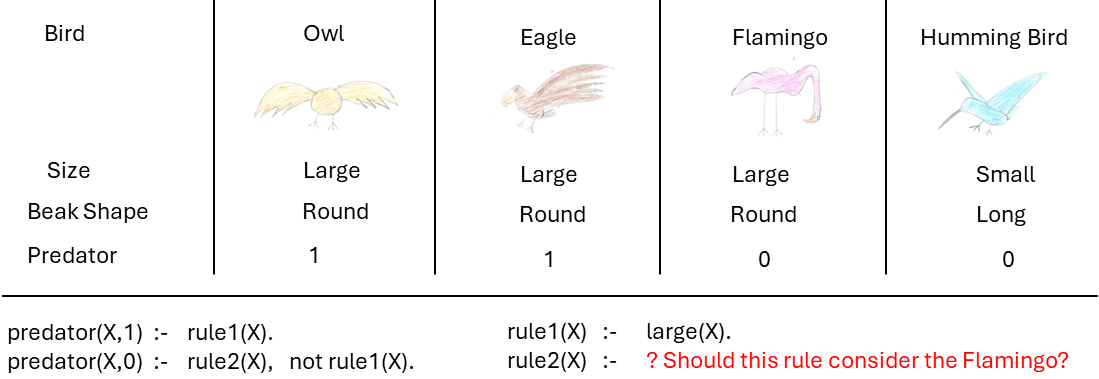}
\caption{\small This toy example illustrates the difference between the FOLD-RM and CON-FOLD core algorithms. Both produce rules of the form shown. CON-FOLD would not consider the Flamingo as part of the data to fit when generating rule 2. FOLD-RM would consider the Flamingo. Note that in many cases both algorithms would generate an abnormal rule \code{ab(X) :- flamingo(X)}, preventing the Flamingo from being covered by the first rule. In this case both FOLD-RM and CON-FOLD would include the Flamingo. When harsh pruning occurs and there are few abnormal rules, this subtle change becomes noticable.}
\label{fig:algorithmchange}
\end{figure*}

\begin{algorithm}[H]
\caption{CON-FOLD}
\label{alg:CON-FOLD}
\begin{pseudo}*
  \hd{CON-FOLD}(X, t) \\*&
  Input: \textnormal{$X$ training examples, $t$ threshold} \\*&
  Output: \textnormal{program $P$ for classifying $X$}\\[1ex]
  $P \gets \emptyset$ \ct{Result} \\
  while $X \neq \emptyset$ do \\+
    $l \gets \pr{most}(X)$ \\
    $X_p, X_n \gets \pr{split\_by\_literal}(X,l)$\\
    $r, R \gets \pr{learn\_rule}(l,X_p, X_n)$\\
    $R \gets \pr{evaluate\_exceptions}(r, R, X_p, X_n, t)$\\
    $X_{\id{fn}} \gets \{d \in X_p \mid R \cup \{r\} \cup{\id{features}(d)} \not\models l\}$\\
    if $|X_{\id{fn}}| = |X_p|$ then break \ct{End if rule does not correctly classify any examples } \\
    $X_{\id{tn}} \gets \{d \in X_n \mid R \cup \{r\} \cup \id{features}(d) \not\models l\}$\\
    $X \gets X_{\id{fn}} \cup X_{\id{tn}}$\\
    $c \gets \id{conf}(R \cup \{r\}, X_p, X_n)$\\
    $P \gets P \cup R \cup \{c::r\}$ \\-
return $P$
  \end{pseudo}
\end{algorithm}
The CON-FOLD algorithm (Algorithm \ref{alg:CON-FOLD}) closely follows the
presentation of FOLD-RM in \cite{Wang2022FOLD-RM}; see there for definitions of
\pr{split\_by\_literal}, \pr{learn\_rule} (slightly adapted) and \pr{most}.
On line 11, \define{$\id{conf}(P, X_p, X_n)$} computes the Wilson score $p$ as in~\eqref{eqn:confidence},
letting
\begin{flalign*}
n_p & = |\{d \in X_p \mid P \cup \id{features}(d) \models \ell\}|\text{, }\\
n & = n_p + |\{d \in X_n \mid P \cup \id{features}(d) \models \ell\}|
\end{flalign*}
and $\ell$ the target class as an atom.
Note that line 6 is only used if pruning. Other than this, the only difference between
CON-FOLD and FOLD-RM are lines 10 and 11. In our terminology, FOLD-RM includes in the
updated set $X$ the full set $X_n$ instead of $X_{\id{tn}}$. The consequence of this
difference is highlighted in Figure~\ref{fig:algorithmchange}.

\begin{theorem}
    CON-FOLD algorithm always terminates on any set of finite examples.
\end{theorem}
\begin{proof}
Each pass through the while loop produces a rule that will either successfully classify at
least one example or not. If no examples are successfully classified, the algorithm
terminates immediately (lines 8-9). Otherwise the examples classified are removed from the
set (line 11) strictly decreasing its size. Since the set is finite and each cycle removes
at least one element, it will become empty eventually and the loop will terminate on that
condition. \hfill\qed
\end{proof}

\noindent FOLD-RM has a complexity of $\mathcal{O}(N M^3)$ where N is the number of features and M is the number of examples~\cite{Wang2022FOLD-RM}. 
In the worst case each literal only covers one example and requires an additional $M-1$ literals to exclude the remaining data. 
Then the pruning algorithm can be called once per rule for each target class, which in the worst case is $M$ times. The complexity of the pruning algorithm is below.

Once confidence values have been assigned to each rule, it is possible to use these values to allow for pruning and prevent overfitting.
In the following we introduce two such pruning methods, \emph{improvement threshold} and \emph{confidence
threshold} pruning.

\vspace{-0.5cm}

\subsection{Improvement Threshold Pruning}
\label{sec:improvement-threshold-pruning}
Improvement Threshold Pruning is designed to stop rules from overfitting data by becoming
unnecessarily `deep'. The rule structure in FOLD allows for exceptions, for
exceptions to exceptions, and then for exceptions to these exceptions and so on. This may
overfit the model to noise in the data very easily. We will refer to any exception
to an exception at any depth as a \textit{sub-exception}. 

In order to avoid this overfitting, each time a rule is added to the model each exception
to the rule is temporarily removed and a new confidence score is calculated. If this
changes the confidence by less than the improvement threshold then this exception is
removed. If an exception is kept then this process is applied to each sub-exception. This
process is repeated until each exception and sub-exception
has been checked.

\begin{algorithm}[ht]
\caption{Evaluate Exceptions}
\label{alg:evaluate}
\begin{pseudo}*
  \hd{evaluate\_exceptions}(r, R, X_p, X_n, t) \\*&
  \hspace*{-1em} Input: \\*&
  \tn{$r$ - rule of form $h_0 \CM l_{r_1}, \vec e_0$ where $l_{r_1}$ \textnormal{is the head of the rule} $r_1 \in R$}\\*&
  \tn{$R$ - set of auxiliary rules for $r$, \textnormal{each of form} $h \CM \vec l, \vec e$ \textnormal{where} $\vec e= \NOT e_1, ..., \NOT e_n $} \\*&
  \tn{\quad where each $e_i$ is the head of a rule $\vec r_{e_i} \in R$}\\*&
  \tn{$X_p$ - positive examples, $X_n$ - negative examples, $t$ - threshold} \\*&
\hspace*{-1em}  Output: \tn{Pruned version of auxiliary rules $R$}\\[1ex]
  return $\pr{ev\_ex\_loop}(r, R, r_1, X_p, X_n, t)$
  \end{pseudo}
  \begin{pseudo}*
  \hd{ev\_ex\_loop}(r, R, r', X_p, X_n, t) \\*&
  \hspace*{-1em} Input: \\*&
  \tn{$r, R, X_p, X_n, t$ as in \pr{evaluate\_exceptions}}\\*&
  \tn{$r'$ - rule in $R$ of form $h \CM \vec l, \vec e$ where $\vec e= \NOT e_1, ..., \NOT e_m$}\\*&
  \tn{\quad and each $e_i$ is the head of a rule $ r_{e_i} \in R$}\\*&
\hspace*{-1em}   Output: \tn{Pruned version of auxiliary rules $R$} \\[1ex]
  $R' \gets R$\\
  for $e_i \in \vec e$ do\\+
  $R_t \gets remove\_rule(R', r_{e_i})$\\
  $c \gets \id{conf}(r, R', X_p, X_n)$\\
  $c_t \gets \id{conf}(r, R_t, X_p, X_n)$\\
  if $c - c_t < t$ then \\+
  $R' \gets R_t$ \ct{Tolerable loss of confidence}\\-
  else \ct{Recurse into exceptions for $e_i$} \\+
    $R' \gets \textnormal{ev\_ex\_loop}(r, R', e_i, X_p, X_n, t)$ \\--
  return $R'$
  \end{pseudo}
\end{algorithm}

The details are formalized in the pseudocode in Algorithm~\ref{alg:evaluate}. In there, 
\define{$\id{remove_rule}(R,r)$} removes the rule $r$ from 
the set $R$ of rules and removes the possibly default-negated atom with the head of $r$ from the bodies of
all rules in $R$. 

\begin{theorem}
    The Evaluate Exceptions algorithm always terminates on any set of finite set of rules, exceptions and data points.
\end{theorem}

\begin{proof}
    On any given rule, each exception is checked once and the algorithm terminates when
    there are no more exceptions to be checked. The depth of recursions and therefore the
    number of times that \pr{evaluate\_exceptions} is called recursively is bound by the
    number of exceptions.  \hfill\qed
\end{proof}

To determine the complexity of this pruning algorithm let $R$ be the total number of exceptions and sub-exceptions and $M=|X_p|+|X_n|$ be the number of examples. Therefore the number of sub-exceptions to any exception is bounded in the worse case by $R-1$ or $\mathcal{O}(R)$.
The calculation of confidence within the loop takes $\mathcal{O}(MR)$ time as it requires comparing each sub-exception to a maximum of $M$ data points.  
The loop must run once for each exception which in the worst case is $\mathcal{O}(R)$. 
Therefore the total complexity of running the pruning algorithm is $\mathcal{O}(MR^2)$.

\subsection{Confidence Threshold Pruning}
\label{sec:confidence-threshold-pruning}
In addition to decreasing the depth of rules it is also desirable to decrease the number
of rules. A \define{confidence threshold} is used to determine whether a rule is worth keeping in
the model or whether it is too uncertain to be useful. If the rule has a confidence value
below the confidence threshold then it is removed. This effectively means that rules could
be removed on two grounds: 
\begin{itemize}
    \item There are insufficient examples in the training data to lead to a high confidence value. 
    \item There may be many examples in the training data which match the rule, however a
      large fraction of the examples are actually counterexamples which go against the
      rule.  
\end{itemize}

\begin{figure*}[ht]
\centering
\includegraphics[width=0.9\textwidth]{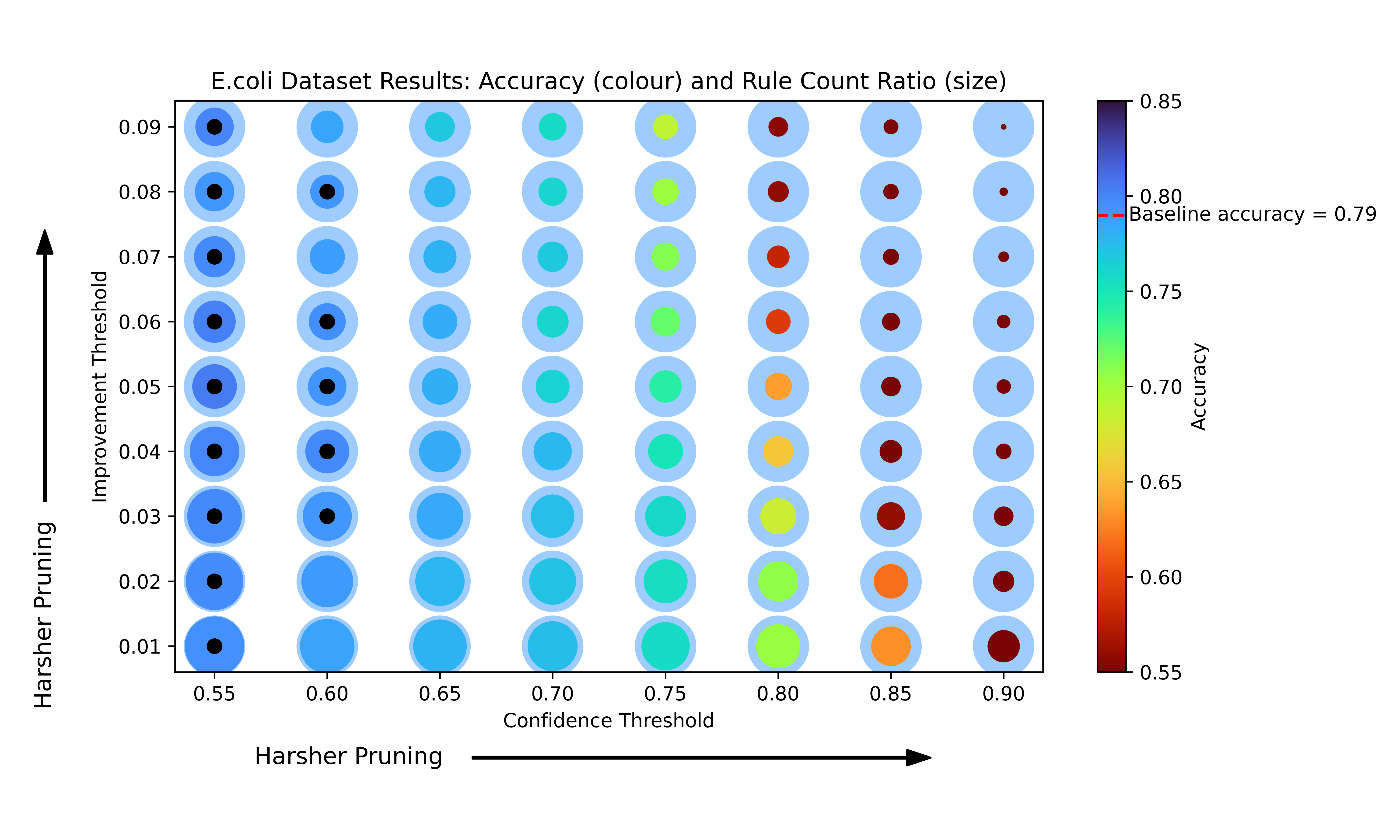}
\caption{\small Scatter plot of the accuracy and number of rules for a ruleset generated by the
  pruning algorithm with different values of the improvement threshold and the confidence
  threshold. Each point has two circles. The background circle displays the number of
  rules and accuracy for no pruning; therefore all background circles are the same. The
  front circle displays the rules and accuracy when pruning is applied. The accuracy is
  indicated by the colour shown in the scale-bar on the right hand side. Pruning
  conditions that are more accurate than the unpruned condition are indicated with a black
  dot in the centre. The number of rules is indicated by the area of the circle (equal
  amount of ink for number of rules), normalised by the number of rules in the unpruned
  case. The results shown for both the accuracy and the number of rules are the average of
  300 trial runs for each test condition.}
\label{fig:ecoli}
\end{figure*}

The two pruning methods introduced above can greatly simplify a model making it more human
interpretable.
However, the pruning of rules reduces the model's ability to fit noise and can lead to an increase
in accuracy. However if rules are pruned to harshly then the model will be
underfit and performance will decrease.
In order to assess these effects we applied CON-FOLD to a sample dataset, the E.coli 
dataset. In our experiments, we varied the values for the two threshold parameters corresponding to the two
pruning methods. This allowed us to assess the performance with respect to accuracy and
to derive recommendations for parameter settings, see Figure \ref{fig:ecoli}.
For this dataset, accuracy is highest for pruning with a low confidence
threshold and a moderate improvement threshold. If the pruning is too harsh this leads to
a significant decrease in performance.  

\section{Inverse Brier Score}
A standard measure of performance in ML is accuracy which for multi-class tasks is defined by:
\begin{equation}
    \text{Accuracy} = \frac{1}{N} \sum^N_{i=1} A(y^*_i,y_i) 
\end{equation}
Where: $A(y^*,y) = 1$ if $y^*=y$ and $0$ otherwise. 

When predictions are made with confidence scores it is possible to use more sophisticated measures of the model's performance. In particular the model should only be given a small reward if it makes a correct prediction with a low confidence, and the model should be punished when making high confidence predictions that are incorrect. We have three desiderata that for such a scoring system:
\begin{enumerate}
    \item It is a proper scoring system: the maximum reward can be gained when the probability value given matches the true probability value.
    \item The scoring system reduces to accuracy when non-probabilistic predictions are made. This allows for the comparison between probabilistic and non-probabilistic models in terms of performance.
    \item The scoring system has an inbuilt mechanism for dealing with no given prediction.
\end{enumerate}

In order to meet all three of these desiderata we propose a variant of the Brier scoring
system and call it Inverse Brier Score (IBS). Brier score (or quadrature score) is often used to evaluate the quality of weather forecasts \cite{Allen2023Conditional, Liu2023Research} and can be obtained with the following formula \cite{Brier1950}:
\begin{equation}
    \text{Brier Score} = \frac{1}{N} \sum^N_{i=1} \sum^K_{k=1} (p_{ki}-y_{ik})^2\text{ ,}
\end{equation}
where $i$ is an index over each data point in a test dataset, $k$ corresponds to a class in the dataset. $N$ is the total number of test examples and $K$ the total number of classes. This is commonly reduced to the following when only making predictions for one class \cite{Murphy1967}:
\begin{equation}
    \text{One Class Brier Score} = \frac{1}{N} \sum^N_{i=1} (p_{i}-y_{i})^2
\end{equation}
We define the IBS as:
\begin{equation}
    \text{IBS} = 1 - \frac{1}{N} \sum^N_{i=1} (p_{i}-y_{i})^2
\end{equation}
 
\begin{theorem}
    Inverse Brier Score is a proper scoring system.
\end{theorem}

\begin{proof}
     It known that the Brier Score is a proper scoring system
     \cite{Murphy1967}. Since the propriety of a scoring system is preserved under linear
     transformations, the IBS is a proper scoring system.  \hfill\qed
\end{proof}

\begin{theorem}
  When definite predictions are made ($p=1$ or $p=0$),
  IBS 
  is equivalent to accuracy.
\end{theorem}

\begin{proof}
    For non-probabilistic decisions, $p_{i}$ is replaced by the prediction
    $y^*$. Therefore $(y_i^*-y_i)^2 = 1-A(y_i^*,y_i)$ and the IBS reduces to the
    definition of accuracy as follws:
    \begin{equation*}
    1 - \frac{1}{N} \sum^N_{i=1} 1-A(y_i^*,y_i) = 1 - \frac{N}{N}+ \frac{1}{N} \sum^N_{i=1} A(y_i^*,y_i)
            = \frac{1}{N} \sum^N_{i=1} A(y_i^*,y_i)\qquad\text{\qed}
    \end{equation*}
\end{proof}

Finally, IBS has a natural way of dealing with no predictions being made. If any class was given a prediction of $p=0.5$, then IBS would be $0.75$ regardless of the class that is chosen. This provides a natural default value if a model refuses to make a prediction due to low confidence, satisfying the third desideratum. Note that this is also important for the FOLD algorithm because it is possible that it will find a test sample which is significantly different to the training examples and may not match any rules.

Therefore, the IBS satisfies all three desiderata. Both IBS and accuracy are used to
evaluate CON-FOLD against other models in Section \ref{sec:results}.
We note that confidence values are regularly used in weather forecasting where Brier score is used as a metric to evaluate their quality. We use this metric to show that FOLD models with confidence scores obtain higher scores. Although Brier Score does not always align with user's expectations \cite{jewson2004problem}, it is a metric which indicates that models with confidence are more interpretable.

\vspace{-0.25cm}

\section{Manual Addition of Rules and Physics Marking}
\label{sec:manual-addition-of-rules}
A key advantage of FOLD over other ML methods is that users are able to incorporate
background knowledge about the domain in the form of rules. In addition to fixed
background knowledge we include modifiable initial knowledge. Initial knowledge can be
provided with or without confidence values.
During training CON-FOLD can add confidence values to provided rules, prune exceptions and even
prune these rules if they do not match the dataset.  

In order for rules to be added to a FOLD model they must be admissible rules as defined below, with optional confidence values. Admissible rules obey the following conditions:
\begin{itemize}
    \item Every head must have a predicate of the value to be decided.
    \item Each body must consist only of predicates of values corresponding to features.
    \item Bodies can be Boolean combinations literals.
    \item The $<$, $>$, $\leq$, $\geq$, $=$ and $\neq$ operators are allowed for comparison of numeric variables.
    \item For categorical variables only $=$ and $\neq$ are allowed.
\end{itemize}
Given formulas of this form are translated into logic programs. Stratified default negation is in place to ensure that rule evaluation is done sequentially, mirroring a decision list structure. Therefore the rules are in a hierarchical structure and the order in which the initial/background knowledge is added can influence model output in the case of overlapping rule bodies. 

Inclusion of background and initial knowledge can be very helpful in cases where only very
limited training data is available. An example of such a problem domain is grading a students'
responses to physics problems. Usually, background domain
knowledge is easily available in the form of well defined rules for how responses should be scored.

We evaluated this idea with data from the 2023 Australian Physics Olympiad provided by
Australian Science Innovations. The data included 1525 student responses to 38 Australian
Physics Olympiad questions and the grades awarded for each of these responses.
We chose to mark the first question of the exam because most students attempted it and the answer is simply a number with units and direction. This problem was also favourable because the marking scheme was simple with only three possible marks, $0$, $0.5$ and $1$. Responses to this question were typed so there was no need to for Optical Character Recognition (OCR) to interpret hand written work. An example of a rule used for marking is:
\begin{verbatim}
  grade(1,X) :- rule1(X).
  rule1(X) :- correct_number(X), correct_unit(X).
\end{verbatim}

\noindent In order to apply the marking scheme, relevant features would need to be
extracted from the student responses. Feature extraction is an active areas of research in
Natural Language Processing \cite{Qi2024English}. Many approaches focus on extracting
information and relationships about 
entities from large quantities of text and
can extract 
large numbers of features \cite{Vicient2013Automatic,
  Carenini2005Extracting}. 
The tools which use term frequency can struggle with short answers \cite{Liu2018Text} especially if they contain large numbers of symbols and numbers, making them inappropriate for feature extraction for grading physics papers. 

In our work we use SpaCy \cite{spacy} for Part of Speech (POS) tagging to extract noun
phrases and numbers to be used as keywords. The presence or absence of these keywords is then one hot
encoded for each piece of text to create a set of features. This method alone was not
sufficient to extract sufficiently sophisticated features that would allow the marking
scheme to be implemented. Therefore we also used regular expressions to extract the
required features for the marking scheme. This required significant
customisation to match the wide variety of notations used by students.

\section{Results}
\label{sec:results}
We compare the CON-FOLD algorithm to XGBoost~\cite{XGBoost}, a standard ML method, to
FOLD-RM, and FOLD-SE, which is currently the state of the art FOLD algorithm. The results
can be found in Table 1.  The experiments use datasets from the UCI Machine Learning
Repository \cite{UCIdataset}. The comparison uses 30 repeat trails where a random 80\% of
the data is selected for training and the remaining 20\% is used for testing.  

\noindent The hyperparameters for all XGBoost experiments are the defaults in the existing
Python implementation\footnote{max\_depth $= 6$, learning\_rate $= 0.3$, n\_estimators
  $= 100$, objective $=$ binary:logistic and scale\_pos\_weight $= 1$}.  In all FOLD-RM
and CON-FOLD experiments, the ratio is set to $0.5$ (default in FOLD-RM).

\noindent\begin{minipage}[t]{0.45\textwidth}
The hyperparameters for the pruned CON-FOLD algorithm are included with the results.  FOLD-SE
was not included in the experiments as the implementation is not publicly available, but
values from Wang et. al. 2023 \cite{Wang2023FOLD-SE} are included. We note that the
accuracy and number of rules for XGBoost and FOLD-RM are very similar to the results given
in the 2023 study and are confident that the experiments are equivalent. The times
measured are wall time measured when the result is run on a PC with an Intel Core
i9-13900K CPU with 64GB of RAM.
Note that when only small amounts of training data were used we ensured
that at least one example of each class was included in the training data; we will refer
to this as stratified training data.
\end{minipage}
\quad
\begin{minipage}[t]{0.48\textwidth}
  \vspace{-0.80cm}
\begin{figure}[H]
\includegraphics[width=\textwidth]{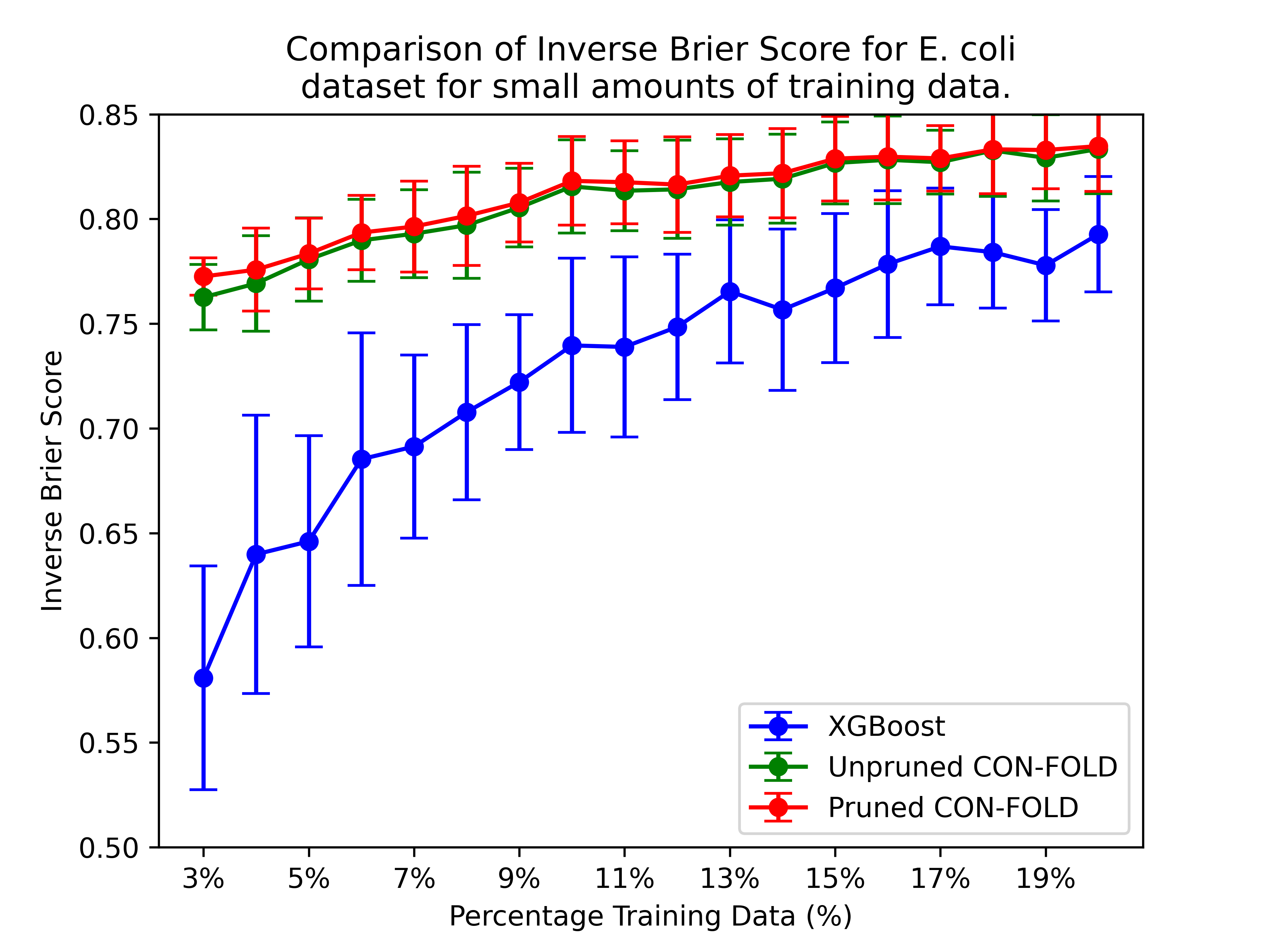}
 \caption{Plot of IBS against percentage of data included in the stratified training data for the E.coli UCI dataset. Thirty trials for each condition were performed and error bars indicate one standard deviation across the trials. Pruned CON-FOLD used a confidence threshold of $0.65$ and a pruning threshold of $0.07$.} 
   \label{fig:LowBrierScore}
\end{figure}
\FloatBarrier
\end{minipage}

\vspace{0.25cm}

\noindent  Figure \ref{fig:LowBrierScore} demonstrates how the
amount of training data impacts the Inverse Brier Score for XGBoost and the CON-FOLD Algorithm with and
without pruning.

\begin{table*}[h!]
  \centering
\tiny
\begin{tabular}{cccccccccccccccccc}
\toprule
\multicolumn{1}{c|}{Dataset} & \multicolumn{2}{c|}{XGBoost} & \multicolumn{4}{c|}{FOLD-RM} & \multicolumn{6}{c|}{CON-FOLD with pruning} & \multicolumn{3}{c}{FOLD-SE} \\
\midrule
& Time & Acc & Time & Acc & Rules & Preds & $t_{con}$ & $t_{imp}$ & Time & Acc & Rules & Preds & Acc & Rules & Preds \\
\midrule
\multirow{2}{*}{Wine} & \multirow{2}{*}{0.15s} & 0.96 & \multirow{2}{*}{0.02s} & 0.94 & 7.3 & 7.3 & \multirow{2}{*}{0.55} & \multirow{2}{*}{0.02} & \multirow{2}{*}{0.02s} & 0.93 & 6.8 & 7.0 & \multirow{2}{*}{0.95} & \multirow{2}{*}{6.5} & \multirow{2}{*}{7.6} \\
 &  & $\pm$0.03 &  & $\pm$0.03 & $\pm$0.6 & $\pm$0.9 &  &  &  & $\pm$0.03 & $\pm$0.8 & $\pm$0.9 &  &  &  \\
\midrule
\multirow{2}{*}{Ecoli} & \multirow{2}{*}{0.67s} & 0.84 & \multirow{2}{*}{0.04s} & 0.79 & 39.2 & 46 & \multirow{2}{*}{0.65} & \multirow{2}{*}{0.08} & \multirow{2}{*}{0.04s} & 0.80 & 12.2 & 16 & \multirow{2}{*}{0.80} & \multirow{2}{*}{24} & \multirow{2}{*}{45} \\
 &  & $\pm$0.04 &  & $\pm$0.04 & $\pm$4.6 & $\pm$8 &  &  &  & $\pm$0.04 & $\pm$1.8 & $\pm$5 &  &  &  \\
\midrule
Weight & \multirow{2}{*}{9.2s} & 1.0 & \multirow{2}{*}{1.7s} & 0.999 & 14.4 & 16.6 & \multirow{2}{*}{0.90} & \multirow{2}{*}{0.02} & \multirow{2}{*}{1.7s} & 0.990 & 11.3 & 12.4 & \multirow{2}{*}{1.0} & \multirow{2}{*}{7.0} & \multirow{2}{*}{11} \\
Lifting &  & $\pm$0.0 &  & $\pm$0.001 & $\pm$1.2 & $\pm$1.3 &  &  &  & $\pm$0.005 & $\pm$0.8 & $\pm$1.1 &  &  &  \\
\midrule
Wall & \multirow{2}{*}{0.47s} & 0.996 & \multirow{2}{*}{2.0s} & 0.993 & 30.1 & 41 & \multirow{2}{*}{0.65} & \multirow{2}{*}{0.01} & \multirow{2}{*}{2.6s} & 0.988 & 20.0 & 25 & \multirow{2}{*}{0.99} & \multirow{2}{*}{7.1} & \multirow{2}{*}{16} \\
Robot &  & $\pm$0.002 &  & $\pm$0.003 & $\pm$2.4 & $\pm$4 &  &  &  & $\pm$0.003 & $\pm$1.8 & $\pm$3 &  &  &  \\
\midrule
Page & \multirow{2}{*}{0.75s} & 0.972 & \multirow{2}{*}{0.93s} & 0.967 & 65.1 & 112 & \multirow{2}{*}{0.70} & \multirow{2}{*}{0.09} & \multirow{2}{*}{1.9s} & 0.94 & 4.1 & 4.4 & \multirow{2}{*}{0.96} & \multirow{2}{*}{8.5} & \multirow{2}{*}{15} \\
Blocks &  & $\pm$0.004 &  & $\pm$0.005 & $\pm$9.5 & $\pm$26 &  &  &  & $\pm$0.01 & $\pm$1.2 & $\pm$2.0 &  &  &  \\
\midrule
\multirow{2}{*}{Nursery} & \multirow{2}{*}{0.68 s} & 0.999 & \multirow{2}{*}{0.90s} & 0.96 & 71.4 & 26 & \multirow{2}{*}{0.55} & \multirow{2}{*}{0.04} & \multirow{2}{*}{1.9s} & 0.92 & 23 & 15.2 & \multirow{2}{*}{0.92} & \multirow{2}{*}{18} & \multirow{2}{*}{40} \\
 &  & $\pm$0.001 &  & $\pm$0.01 & $\pm$8.0 & $\pm$3 &  &  &  & $\pm$0.01 & $\pm$3 & $\pm$2.5 &  &  &  \\
\midrule
Dry & \multirow{2}{*}{1.3s} & 0.928 & \multirow{2}{*}{9.6s} & 0.911 & 186 & 303 & \multirow{2}{*}{0.65} & \multirow{2}{*}{0.01} & \multirow{2}{*}{14.3s} & 0.90 & 63 & 106 & \multirow{2}{*}{0.90} & \multirow{2}{*}{25} & \multirow{2}{*}{31} \\
Bean &  & $\pm$0.004 &  & $\pm$0.005 & $\pm$16 & $\pm$37 &  &  &  & $\pm$0.01 & $\pm$19 & $\pm$36 &  &  &  \\
\bottomrule
\end{tabular}
 \caption{\small Accuracy, Runtime and Number of Rules and Predicates for different methods and different benchmarks from the UCI repository. Pruning hyperparameters are provided for the CON-FOLD algorithm. The uncertainty values provided are the standard deviations from 30 trial runs. FOLD-SE results are taken from Wang et. al. 2023~\cite{Wang2023FOLD-SE}.
 }
\end{table*}



\noindent One particular use case where high levels of explainability is required is marking
students' responses to exam questions. When artificial intelligence or ML is applied to
automated marking, it is desirable for a minimal training data set. This reduces the cost
to mark initial examples to use as training data for the models. Therefore we tested each
model on very small training data sets with the first having just three examples of
student work ($0.2\%$ of the data set).  

\begin{figure*}[h!]
\centering
\textbf{Plots of Inverse Brier Score against amounts of training data for grading student
  responses to an Australian Physics Olympiad problem with and without manual feature
  extraction} 

\begin{subfigure}{0.49\textwidth}
  \centering
  \includegraphics[width=\linewidth]{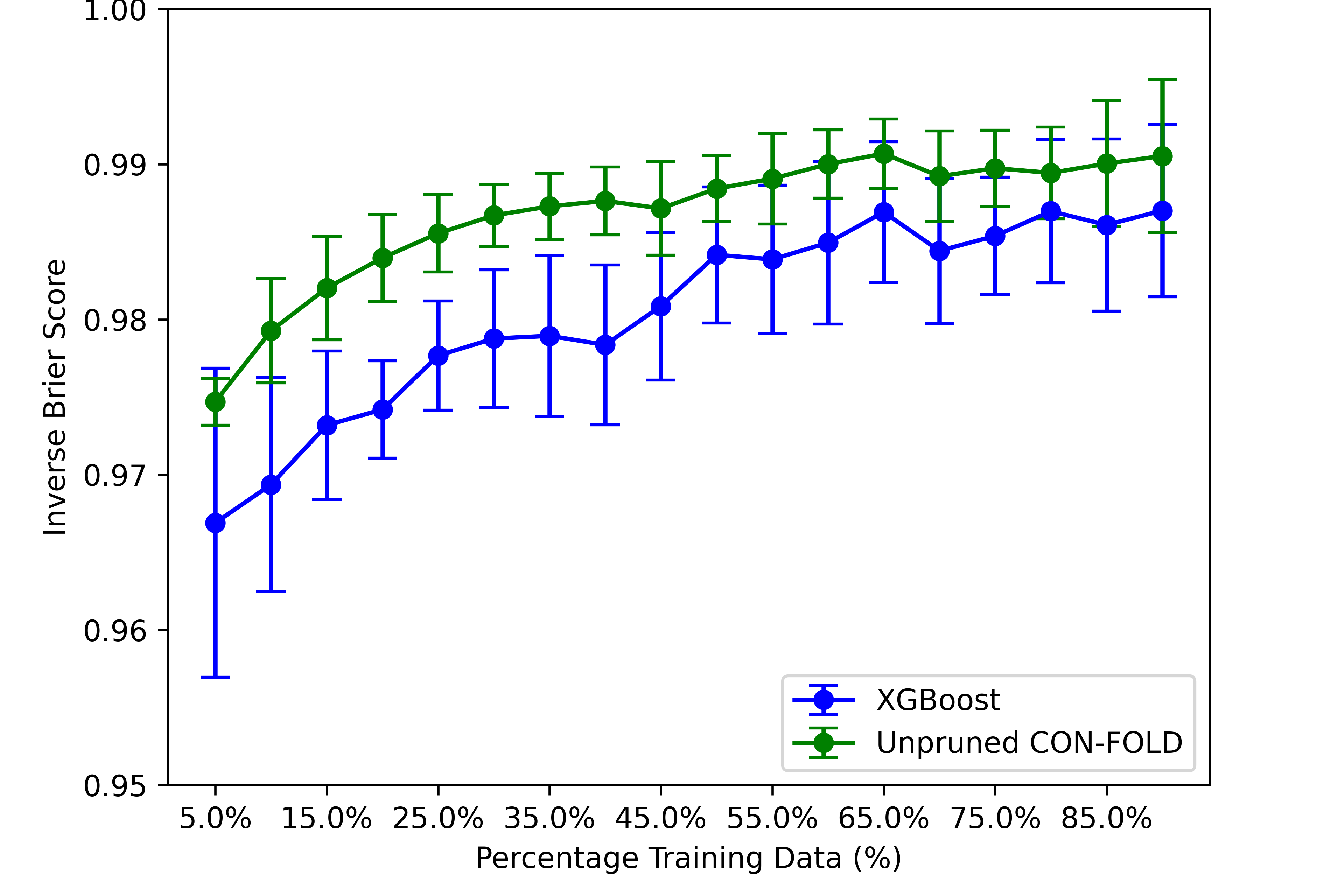}
  \caption{}
  \label{fig:sub1}
\end{subfigure}%
\hspace{0.01\textwidth}
\begin{subfigure}{0.49\textwidth}
  \centering
  \includegraphics[width=\linewidth]{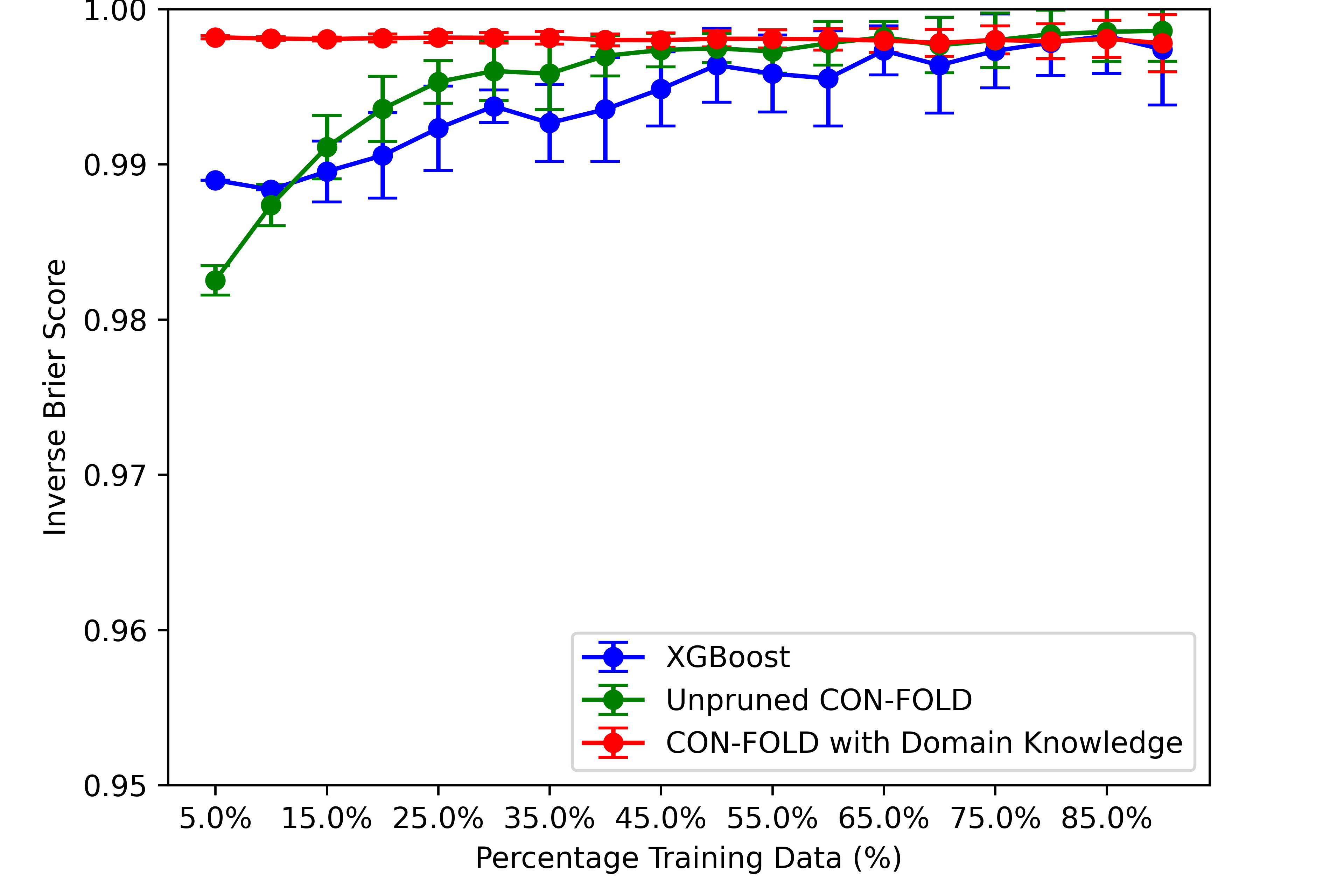}
  \caption{}
  \label{fig:sub2}
\end{subfigure}

\begin{subfigure}{0.49\textwidth}
  \centering
  \includegraphics[width=\linewidth]{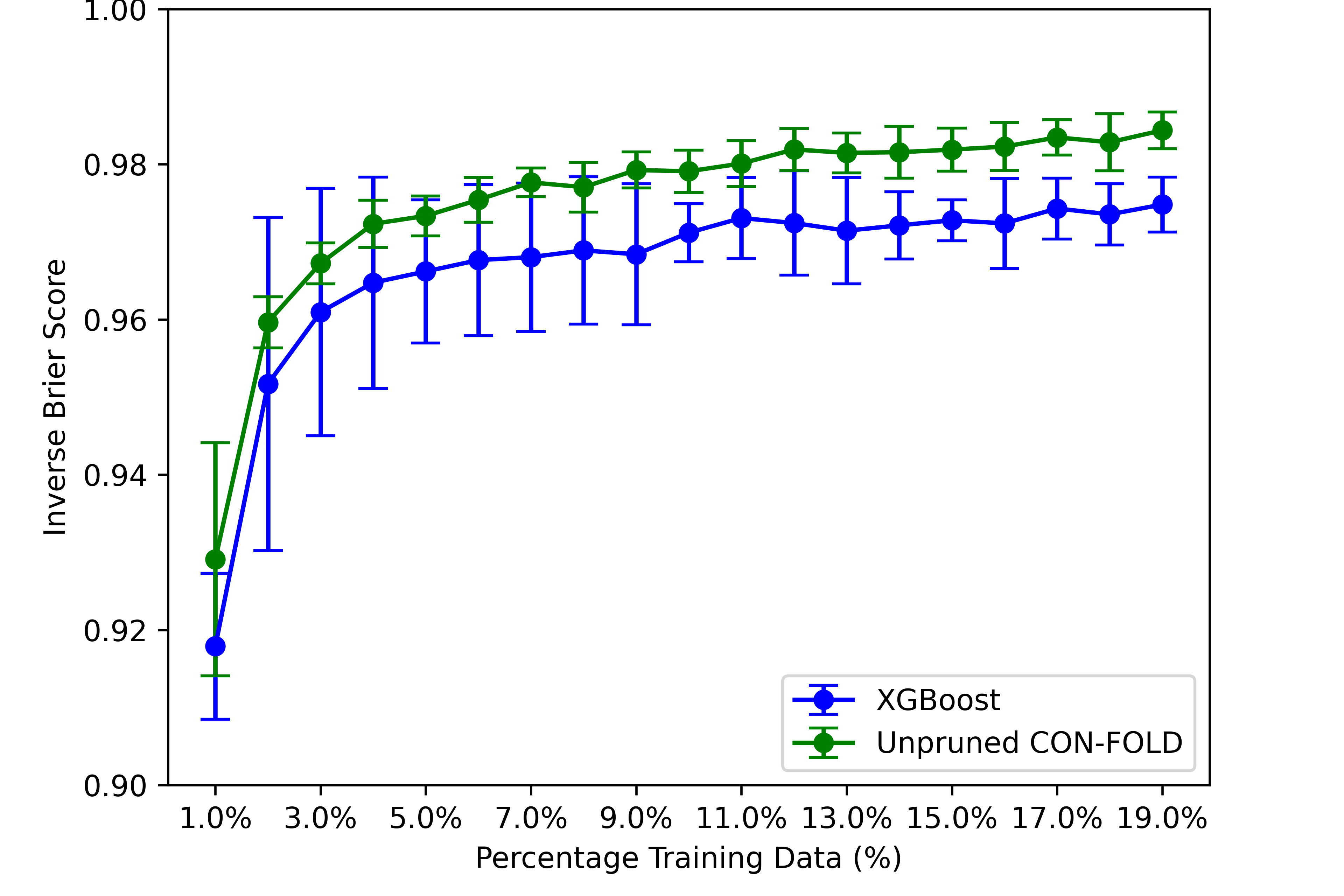} 
  \caption{}
  \label{fig:sub3}
\end{subfigure}%
\hspace{0.01\textwidth}
\begin{subfigure}{0.49\textwidth}
  \centering
  \includegraphics[width=\linewidth]{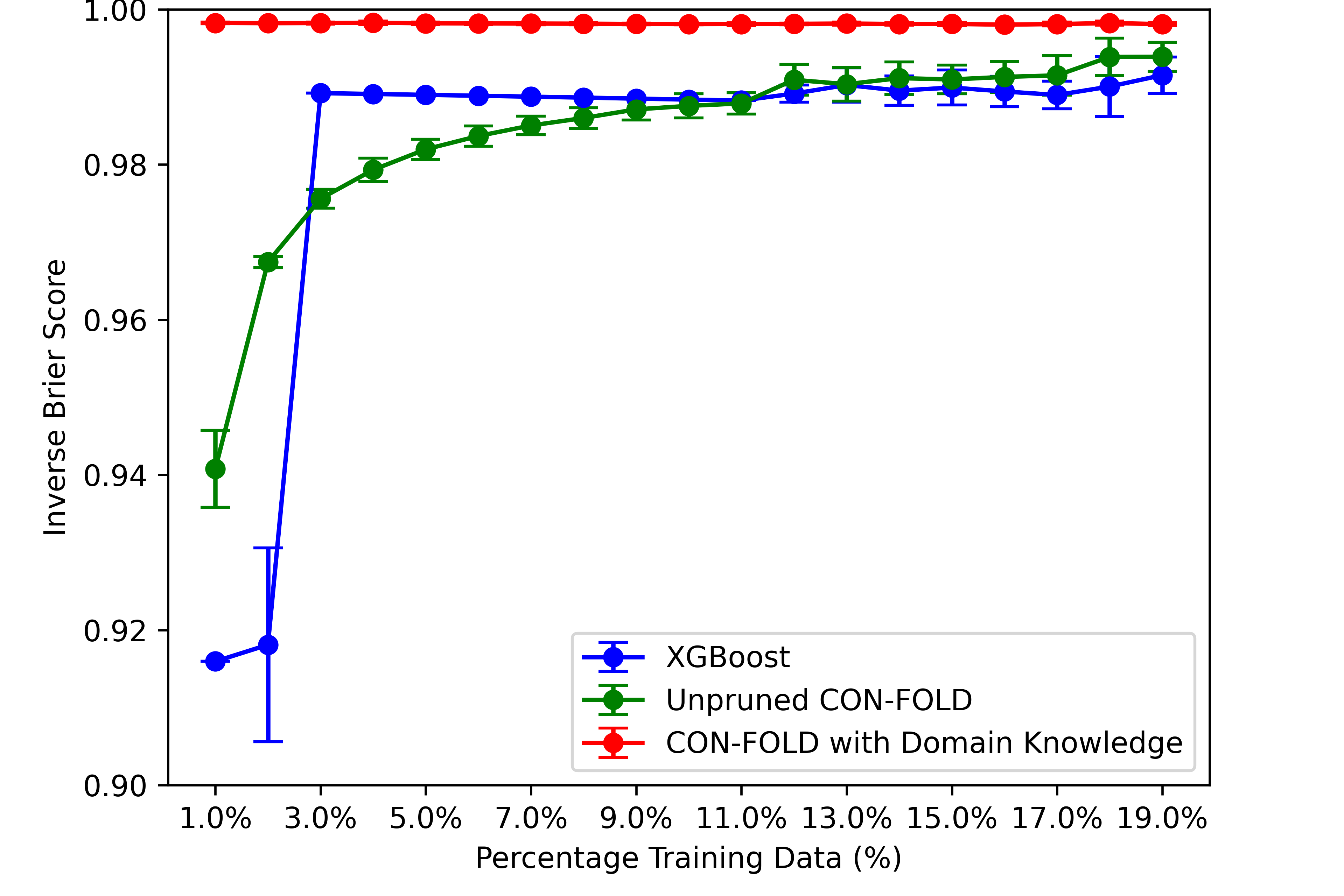}
  \caption{}
  \label{fig:sub4}
\end{subfigure}

\begin{subfigure}{0.49\textwidth}
  \centering
  \includegraphics[width=\linewidth]{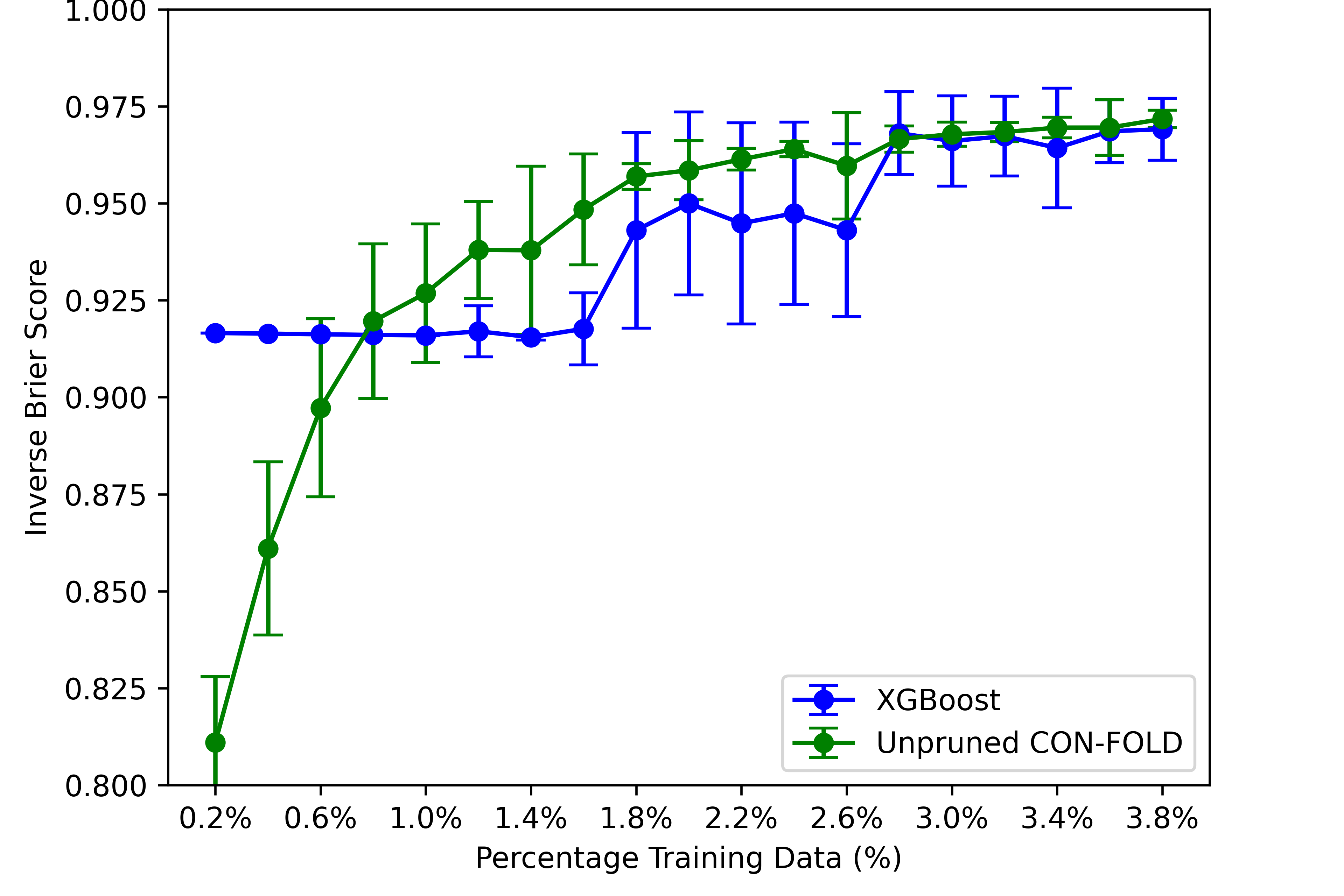} 
  \caption{}
  \label{fig:sub5}
\end{subfigure}%
\hspace{0.01\textwidth}
\begin{subfigure}{0.49\textwidth}
  \centering
  \includegraphics[width=\linewidth]{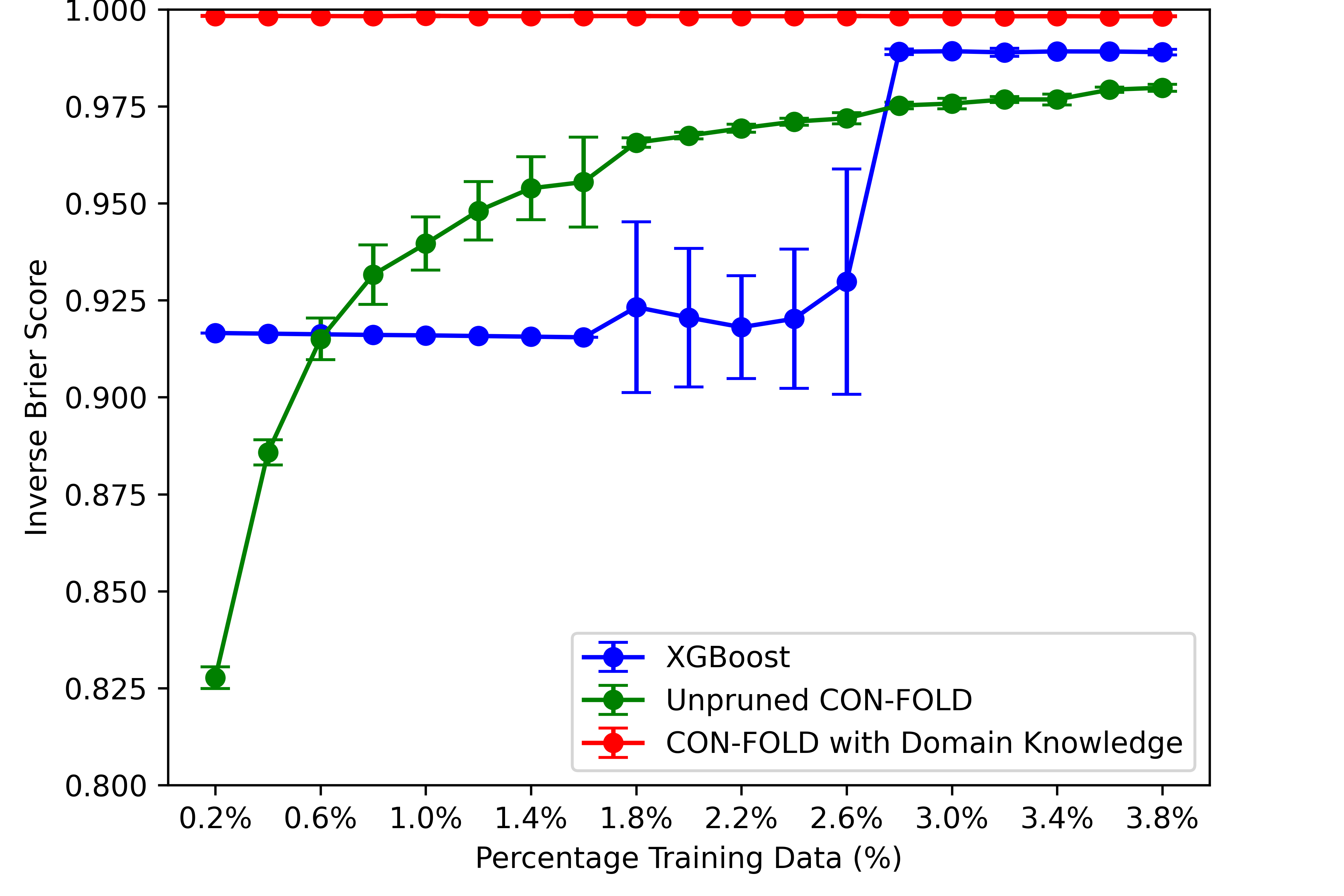}
  \caption{}
  \label{fig:sub6}
\end{subfigure}

\caption{\small Each of the plots shows the performance of models using the Inverse Brier Score
  metric with different amounts of training data. Plots a and b show the regimes where
  large amounts of training data are available while plots e and f explore model
  performance with very small amounts of training data available. Plots a, c and e use
  automatic feature extraction, while plots b, d and f use manual feature extraction using
  regular expressions which allows for domain knowledge in the form of a marking scheme to
  be included. The total number of student responses was $n=1525$.} 
\label{fig:physicsresults}

\end{figure*}

Three different models were tested. The first two were XGBoost and the unpruned CON-FOLD
algorithm. The third model tested was the CON-FOLD algorithm with the marking scheme given
as domain knowledge before training. Each rule from the marking scheme was given a
confidence of 0.99.  
Thirty trials were conducted with randomly generated stratified data ranging from $0.2\%$
to $90\%$ of the data used for training. The results of this experiment can be seen in
Figure \ref{fig:physicsresults}. The average is represented by the points on the graph and
the standard deviation from the thirty trials is given as error bars. 

\vspace{-0.5cm}

\section{Discussion}
The runtime for XGBoost on the weight lifting data in Table 1 appears
anomalously long. This has been observed previously \cite{Wang2023FOLD-SE} and can be
attributed to the large number of features ($m=155$) in the dataset which is an order of
magnitude greater than the others. This result confirms that both CON-FOLD and the pruning
algorithm scales well with the number of input features. 

Table 1 shows that the pruning algorithm reduces the number of rules compared to the
FOLD-RM algorithm with a small decrease in accuracy. The main advantage of decreasing the
number of rules is that it makes the results more interpretable to humans; as having  hundreds of rules becomes quite difficult for a human to follow.  Furthermore a smaller set of rules reduces the inference time of the model.
The FOLD-SE
algorithm produces a smaller number of rules and higher accuracy for most datasets. We
note that the CON-FOLD pruning technique and the use of Gini-Impurity from FOLD-SE are not
mutually exclusive and applying both may result in even even more concise results while
maintaining performance. 

Our experiment in Figure \ref{fig:LowBrierScore} shows that for small
amounts of stratified training data, the ability to put confidence values on predictions
gives CON-FOLD a significant advantage over XGBoost. We also note that pruning gives
a very slight advantage in cases of very small training data sets. We attribute this to
the prevention of the model from overestimating confidence 
from only a small number of examples. 

For the physics marking dataset, the rules created from the marking scheme align almost
perfectly with the scores that students were actually awarded, as expected. This results
in very strong performance even when there is very little training data. We note that the
unpruned CON-FOLD algorithm gradually increases its IBS as the amount of training data
increases. We attribute this to a combination of increasing confidence in rules that were
learned and being able to learn more complex rules from larger training data sets. 

For the XGBoost experiments with features generated by regular expressions in Figure
\ref{fig:sub6}, we note that the algorithm is able to score approximately 0.92
consistently until $1.8\%$ of the training data is reached. Then the model's performance
seems to vary wildly between trials before jumping to an accuracy of 0.99 at $2.8\%$. We
attribute the instability as sensitivity to specific training examples being included in
its training data. Once $2.8\%$ of the data is included in training this seems to settle
as this is a sufficient amount that the required examples reliably fall into the training
data set. A similar pattern is also present for Figure \ref{fig:sub3}.

For small amounts of training data, IBS of the CON-FOLD algorithm is mostly independent of
whether regular expression features were included or not. However in the regime of large
amounts of training data shown in Figure \ref{fig:sub1} and Figure \ref{fig:sub4},
manually extracted features make a small but significant improvement in the performance of
both XGBoost and CON-FOLD. This has implications for the use of automated feature POS
tagging tools when being used for feature extraction for automated marking. Regular
expressions for feature extraction allow for more accurate results and for the
implementation of background rules, but this comes at significant development costs. 

\section{Conclusion and Future Work}
We have introduced confidence values that allow users to know the probability that a rule from a FOLD model will be correct when applied to a dataset. This removes the illusion of certainty when using rules created by the FOLD algorithm. We introduce a pruning algorithm that can use these confidence values to decrease the number and depth of rules. The pruning algorithm allows for the number of rules to be significantly decreased with a small impact on performance, however it is not as effective as the use of the Gini Impurity methods in FOLD-SE. 

Inverse Brier score is a metric that can be used to reward accurate forecasting of probabilities of rules while maintaining compatibility with non-probabilistic models by reducing to accuracy in the case of non probabilistic predictions.

CON-FOLD allows for inclusion of background and initial knowledge into FOLD models. We use
the marking of short answer physics exams as a 
potential use case to
demonstrate the effectiveness of incorporating readily-available domain knowledge in the
form of a marking scheme. With this background knowledge, the CON-FOLD model's performance
is significantly improved and out performs XGBoost, especially in presence of small amounts of
training data.

Besides improvements to the FOLD algorithm, an area for future work is feature extraction
from short snippets of free-form text. NLP feature extraction tools were not able to
capture the required features to implement a marking scheme, but otherwise were able to
extract enough features to allow for accuracy and Inverse Brier Score of over $99\%$ in
the regime of large amounts of training data.
Optical Character Recognition could allow for the grading of
hand-written responses to be explored.  As a final
suggestion for future research, more advanced NLP tools such as Large Language Models
(LLMs) could be used to allow for automated extraction of numbers and units from text.

\paragraph{\textbf{Acknowledgements}}
The authors would like to thank Australian Science Innovations for access to data from the 2023 Australian Physics Olympiad. 
This research was supported by a scholarship from CSIRO's Data61.
The ethical aspects of this research have been approved by the ANU Human Research Ethics Committee (Protocol 2023/1362).
All code can be accessed on Git-Hub at
\noindent \url{https://github.com/lachlanmcg123/CONFOLD}. We thank Daniel Smith for
helpful comments.



\end{document}